\documentclass{article}

\PassOptionsToPackage{numbers, compress}{natbib}

\usepackage[final]{neurips_2019}

\usepackage[utf8]{inputenc} 
\usepackage[T1]{fontenc}    
\usepackage{hyperref}       
\usepackage{url}            
\usepackage{booktabs}       
\usepackage{amsfonts}       
\usepackage{nicefrac}       
\usepackage{microtype}      
\usepackage{graphicx} 
\usepackage{adjustbox}
\usepackage{tabularx}
\usepackage{subcaption} 
\usepackage{float}

\title{Towards Personalized Dialog Policies for Conversational Skill Discovery}

\author{Maryam Fazel-Zarandi,  \\
Amazon Alexa, USA\\
 {\tt fazelzar@amazon.com}\\
\\\And
 Sampat Biswas \\
Amazon Alexa, USA\\
 {\tt sampab@amazon.com} \\
 \\\And
 Ryan Summers \\
Amazon Alexa, USA\\
 {\tt ryansum@amazon.com} \\ 
\\\AND
 Ahmed Elmalt \\
Amazon Alexa, USA\\
 {\tt elmalt@amazon.com} \\
\\\And
 Andy McCraw \\
Amazon Alexa, USA\\
 {\tt andmccra@amazon.com} \\
\\\And
 Michael McPhilips \\
Amazon Alexa, USA\\
 {\tt mcphim@amazon.com} \\
 \\\And
John Peach \\
Amazon Alexa, USA\\
 {\tt jpeach@amazon.com} \\}

\begin{document}

\maketitle

\begin{abstract}
Many businesses and consumers are extending the capabilities of voice-based services such as 
Amazon Alexa, Google Home, Microsoft Cortana, and Apple Siri to create custom voice experiences (also known as skills). 
As the number of these experiences increases, 
a key problem is the discovery of skills that can be used to address a user's request. 
In this paper, we focus on conversational skill discovery and present a conversational agent which engages in a dialog with users to help them find the skills that fulfill their needs. 
To this end, we start with a rule-based agent and improve it by using reinforcement learning.
In this way, we enable the agent to adapt to different user attributes and conversational styles as it interacts with users.
We evaluate our approach in a real production setting by deploying the agent to interact with real users, and show the effectiveness of the conversational agent in helping users find the skills that serve their request.
\end{abstract}

\section{Introduction}

Modern speech-based assistants, such as Amazon Alexa, Google Home, Microsoft Cortana, and Apple Siri, 
enable users to complete daily tasks such as shopping, setting reminders, and playing games 
using voice commands.
Such human-like interfaces create a rich experience for users by enabling them to complete many tasks hands- and eyes-free in a conversational manner.
Furthermore, these services offer tools to enable developers and customers to create custom voice experiences (skills) and as a result extend the capabilities of the assistant.
Amazon's Alexa Skills Kit \cite{Kumar2017}, Google's Actions and Microsoft's Cortana Skills Kit are examples of such tools.
As the number of skills (with potentially overlapping functionality) increases,
it becomes more difficult for end users to find the skills that can address their request. 

To mitigate the skill discovery problem, recently researchers have proposed solutions for personalized domain selection and continuous domain adaptation in speech-based assistants \cite{Kim2018,Han2019}. 
Although such solutions help users find skills, in scenarios such as searching for a game where many different skills exist and user's preferences change,
routing the user to a particular experience
would not be satisfactory. 
In such cases, the assistant should initiate a conversation with the user,
making recommendations, asking for preferences, 
and allowing the user to browse through different options.
Similar to other search problems, personalization is important for conversational skill discovery and can be achieved at two levels: 
1) personalization of skill recommendations, and 
2) personalization of the interaction. 
Users have evolving attributes (e.g., first-time vs returning user) and different conversational styles and preferences (e.g., brief vs verbose communication) which affect how they respond to what the agent is proposing and its recommendations.
By personalizing the interaction according to user attributes, conversational styles and preferences, the speech-based assistant can help speed up the conversation process \cite{Mo2018} and increase user satisfaction.
However, existing works are limited with respect to considering user's evolving attributes and diverse multi-aspect preferences \cite{Rafailidis2019}, such as preferences with respect to how the conversational agent interacts with them.

In this paper, we focus on conversational discovery of skills to guide customers from an intent to a specific skill 
or set of skills that can serve their request.
To this end, we start with a rule-based agent and improve it by using reinforcement learning (RL), 
enabling the agent to adapt to different conversational styles as it interacts with users.
In summary, the contributions of this paper are as follows: 
1) We introduce the problem of conversational skill discovery for large-scale virtual assistants. 
2) We describe a solution which enables the assistant to adapt to user's attributes (e.g., first-time user vs returning user) and conversational styles (e.g., brief vs. verbose). 
3) We conduct experiments in a real production setting by deploying the agent to interact with real users in large scale, showing that the personalized policy learned using RL significantly outperforms a one-fits-all rule-based agent in terms of success rate (measured in terms of number of dialogs which result in launching a skill) with significantly shorter dialogs.

\section{Conversational Skill Discovery}

\begin{table}[!b]
\small
 \centering
 \caption{Example of a conversation with the Skill Discovery Agent}
 \begin{adjustbox}{max width=1.0\textwidth}
 \bgroup
 \begin{tabularx}{0.9\linewidth}{lX}
  \toprule
  \multicolumn{1}{c}{\textbf{Speaker}} & \multicolumn{1}{c}{\textbf{Utterance}} \\
  \midrule
  User & Let's play a game \\
  Agent & OK, I know lots of games. Did you want trivia games, history games, or word games? \\
 User & History games \\
 Agent & Alright, history games. Did you want Ultimate History Quiz? \\
 User & No, give me word games \\
 Agent & OK, Word games. Did you want Amazing Word Master Game? \\
 User & What's its rating? \\
 Agent & Amazing Word Master Game has a rating of 3.5. Would you like to play it? \\
 User & Yes \\
 Agent & OK, here's Amazing Word Master Game. (\textit{skill launched}) \\
 \bottomrule
 \end{tabularx}
 \egroup
 \end{adjustbox}
 \label{codi-example}
\end{table}

Conversational skill discovery is the task of initiating a dialog with the user in order to help them find the 
skills that address their needs when interacting with a speech-based assistant.
More specifically, a conversational skill discovery agent receives a natural language input from the user,
understands it using its automatic speech recognition (ASR) and
natural language understanding (NLU) components, and decides how to respond to the user based on user provided and contextual information in order to help the user find the needed skill.
Skills can often be grouped into categories and subcategories based on functionality (e.g., ride-sharing skills or trivia games).
These categories help customers explore with much more specificity and relevance, as such a key functionality of a skill discovery system is to allow users to browse through existing categories.
Additionally, it is important for the agent to be able to adapt to user's conversational styles, 
overtime shifting to more and more personalized conversations with the user.

Table \ref{codi-example} shows an example of a dialog between a user and an agent.
Here, in each turn of the dialog, the user can either ask for a particular category or skill, 
select from the list of recommendations, accept or reject a recommendation, ask for other (sub)categories or skills, 
ask for details or rating of a skill, 
or perform some general action such as asking for help, asking the agent to repeat the previous prompt, 
going over a list of recommendations, going back in the conversation, or asking the agent to stop.
The agent, on the other hand, can suggest a skill, provide information or help, offer a few different types of categories to choose from,
stop the conversation if it is not going well, or launch a selected skill.

\subsection{Problem Formulation}

Conversational skill discovery, similar to other goal-oriented dialog systems, can be formalized as a Markov Decision Process (MDP) \cite{Levin2000}. 
An MDP is a tuple $<\mathcal{S}, \mathcal{A}, \mathcal{P}, \mathcal{R}, \mathcal{\gamma}>$, where $\mathcal{S}$ is the state space, $\mathcal{A}$ is the action space, $\mathcal{P}$ is the transition probability function, $\mathcal{R}$ is the reward function, and $\mathcal{\gamma}$ is the discount factor. 
In this framework, at each time step \textit{t}, 
the agent observes state $s_t \in \mathcal{S}$ and selects action $a_t \in \mathcal{A}$ according to its policy ($\pi: \mathcal{S} \rightarrow \mathcal{A}$). 
After performing the selected action, the agent receives the next state \textit{$s_{t+1}$} and a scalar reward \textit{$r_t$}. 
The trajectory restarts after the agent reaches a terminal state. 
RL solvers have been used to find the optimal dialog policy (e.g., \citeauthor{Singh2002} \citeyear{Singh2002}; \citeauthor{Williams2007} \citeyear{Williams2007}; \citeauthor{Georgila2011} \citeyear{Georgila2011}; \citeauthor{Lee2012} \citeyear{Lee2012}). 
In this context, at each turn the agent acts based on its understanding of what the user said, 
and reward function is modeled in terms of various dimensions of the interaction such as per-interaction user satisfaction, accomplishment of the task, efficiency of interaction, and dialog duration. 
Recently, deep RL has also been applied to the problem of dialog management and has shown improvements over rule-based systems \cite{Cuayahuitl2016,Zhao2016,Fatemi2016,Fazel2017,Liu2018}. 

In this paper, we adopt the above formalism with the goal of training a dialog policy which allows the agent to take actions that maximize its success rate (measured in terms of number of dialogs which result in launching a skill)
while providing a flexible and natural way for the user to navigate throughout various dialog states.
In each turn of the dialog, the agent makes its decisions based on various available information such as
user's intent (e.g., asking for a particular skill),
the category the user has selected,
whether the user is a first-time user, etc.
In order to make the agent adapt to different conversational styles,
when making recommendations, we focus on 
1) whether to recommend skills or categories, 
2) how many skills or categories to recommend, and 
3) what type of metadata to provide to the user. 
Examples of metadata include: popularity, star rating, number of reviews, or a short description of the skill.
The agent can proactively provide metadata to the user at certain points in the experience. 
Depending on user's conversational style, they may prefer brief conversations with the agent (i.e., no metadata),
or verbose with different types of metadata.

An important challenge in using RL for learning dialog policies is creating realistic user simulators 
that can generate natural conversations similar to a human user \cite{Schatzmann2006}, and as such
in previous works researchers have focused on the development of different types of user simulators (e.g., \citeauthor{Eckert1997} \citeyear{Eckert1997}; \citeauthor{Scheffler2002} \citeyear{Scheffler2002}; \citeauthor{Cuayahuitl2005} \citeyear{Cuayahuitl2005}; \citeauthor{Georgila2006} \citeyear{Georgila2006}; \citeauthor{Schatzmann2006} \citeyear{Schatzmann2006}; \citeauthor{ElAsri2016} \citeyear{ElAsri2016}; \citeauthor{XiujunLi2016} \citeyear{XiujunLi2016}). 
We take a data-driven approach to user simulation, and start with a rule-based policy to gather data and then improve the agent by using RL.

\subsection{Rule-based Agent}

The rule-based agent selects from the following actions depending on user's intent in each turn of the dialog: 
1) offering \textit{k} categories ($1 \leq k \leq 5$), 
2) offering \textit{n} skills ($1 \leq n \leq 3$),  
3) offering a skill or asking for category,
4) providing information about skill rating, 
5) providing details about a skill,
6) ending the conversation, and 
7) launching a skill. 
When multiple actions are possible, the rule-based agent randomly selects among them. 
For example, at the beginning of the dialog, the agent randomly selects among different offer actions.
If all skills in a category have been exhausted, the agent will inform the user that no additional skills are available for the selected category.
Furthermore, each action is mapped to a specific prompt template. 
For example, \textit{offering a skill or asking for category} can be mapped to 
"Would you like to launch $<$skill$>$ or try a different type of skill?", 
where the specific skill is provided by a skill recommendation system. 
Additionally, in cases where the agent does not understand what the user has said (e.g., out-of-domain requests), it will first repeat the previous prompt, if user's utterance is again misunderstood, it will give a new prompt, and finally it will stop the conversation.

\subsection{User Simulation}

\begin{figure}[t]
\centering
    \caption{User simulator and conversational agent interaction.}
  		\includegraphics[width=0.85\textwidth]{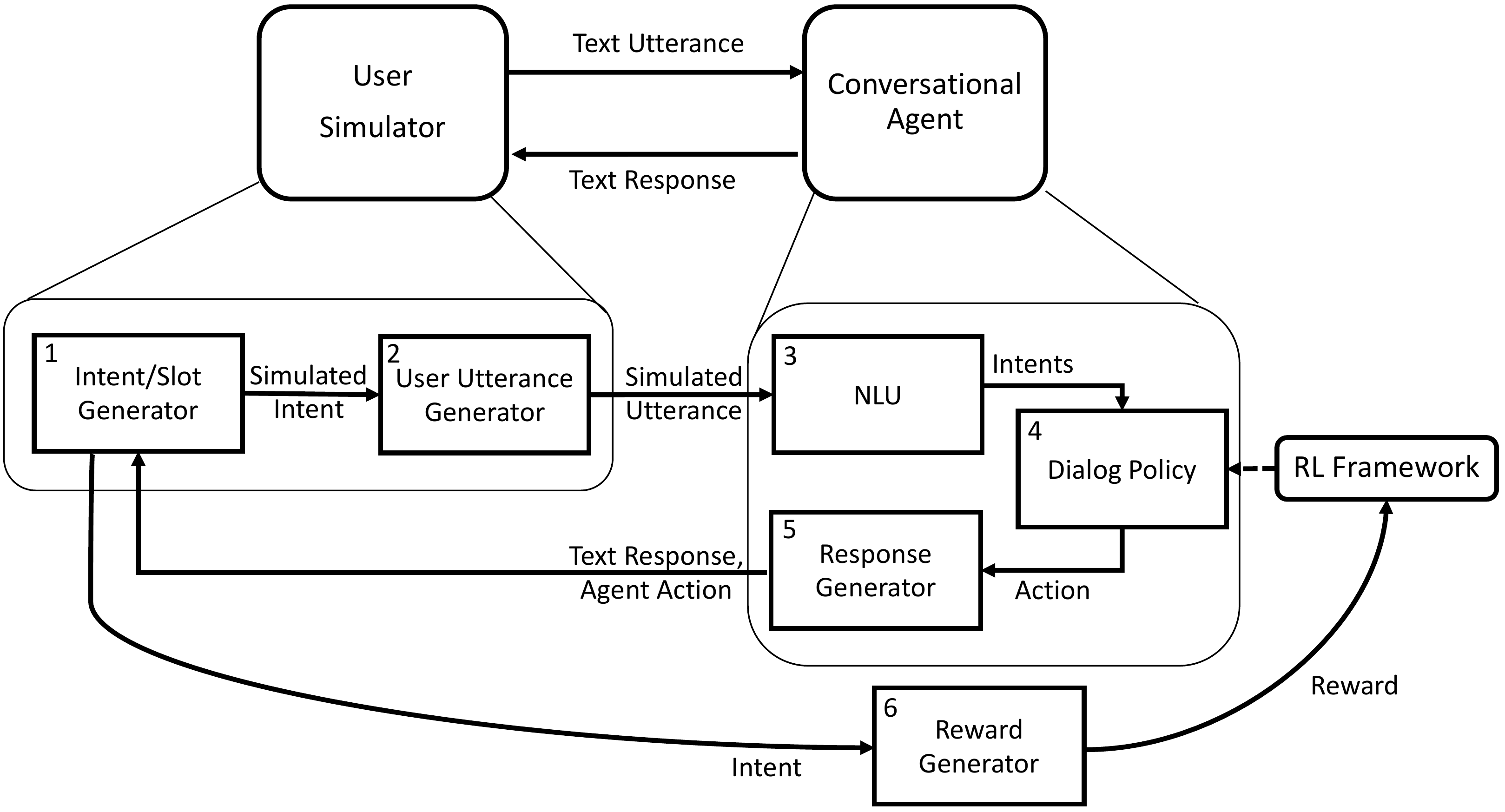}
  \label{simulator}
\end{figure}

We deployed the rule-based system to gather dialogs with users and 
trained a user simulator similar to \citet{Fazel2017} with $180,000$ dialogs with real users. 
Note that the collected dialogs are not annotated and may include understanding errors. Figure \ref{simulator} illustrates the interaction between the user simulator (left) and the conversational agent (right).  
More specifically, the user simulator first generates the next user intent based on dialog context. 
Intent generation is modeled as a language modeling problem.
In this formulation, each possible intent forms a token in the vocabulary, and every training dialog becomes a training intent sequence. For example, the sequence for the conversation in Table \ref{codi-example} is \textit{[Start, CategoryName, CategoryName, GetRating, Yes, End]}.

We used recurrent neural networks with Gated Recurrent Unit (GRU) \cite{Cho2014} to predict the next user intent, 
and used the following for dialog context:
1) previous user intent, 
2) previous agent action, 
3) previous agent prompt, 
4) whether the user is a first-time user, 
5) whether the user has already selected an item (skill or category) from a list,
and 
6) number of user turns so far in the conversation.
The optimal parameters were found using Hyperopt \cite{hyperopt} and the model with lowest perplexity \cite{Serban2016} score was chosen.
Given the predicted intent, the user simulator uniformly samples one utterance from the combination of available templates and user turns in the collected dialogs. 

\subsection{RL-based Agent}

The components used to learn dialog policies using RL are as follows. 

\vspace{2pt}
\noindent
\textbf{State Space \textit{S}}: The input state is composed of 
1) user's intent 
2) previous action the agent took, 
3) previous prompt and metadata it gave the user,
4) the category the user has selected if any,
5) whether the agent has proposed a skill,
6) whether the user is a first-time user, and
7) number of user turns so far in the dialog. 
This set of parameters were selected using a forward feature selection approach
based on the correlation between the new feature and the feature set with the goal of achieving a higher Expected Cumulative Reward (ECR) \cite{Shen2016}.
This set can be augmented with user preferences regarding skills, 
the last skill launched by the user, or
the frequency of skill launches. 

\vspace{2pt}
\noindent
\textbf{Action Space \textit{A}}: We constrain the action space of the agent to a set of composite actions: 
1) offering \textit{k} categories (e.g., offer-one-category, offer-two-category), 
2) offering \textit{n} skills (e.g., offer-one-skill, offer-two-skill),  
3) offering a skill or asking for category (e.g., offer-one-skill-or-category),
4) executing a user request, 
5) ending the conversation, and 
6) launching a skill. 
The \textit{execute} action refers to delivering information such as providing skill ratings or more details about a skill, 
repeating the previous prompt, 
or handling out-of-domain requests. 
At run-time, the RL policy falls back on the rule-based policy for the \textit{execute} action.

\vspace{2pt}
\noindent
\textbf{Reward \textit{R}}: We use a simple reward function based on goal completion, where the environment gives a reward of $+1$ at the end of the dialog if the user launches a skill, and
gives a reward of $-1$ if the user or agent end the dialog. 

\begin{table*}[t]
 \centering
  \caption{State and Action Spaces}
 \begin{tabularx}{1.0\linewidth}{lcl}
 \toprule
   \textbf{Variable} & \textbf{Possible Values} &  \multicolumn{1}{c}{\textbf{Example}}\\
 \midrule
	State - User Intent & $17$ & start, category-name, skill-name, stop, etc. \\
	State - Previous Agent Action & $8$ & offer-one-skill, offer-one-skill-or-category, \\
	& & offer-one-category, offer-three-categories, \\
	&& offer-five-categories, launch-skill, \\
	& & end-session, execute\\
	State - Previous Prompt & $56$ &  first-time-user-offer-three-categories, etc.\\
	State - Previous Metadata & $5$ & no-metadata, short-description, trending, \\
	& &recommended, rating-review\\
	State - Target Category & $191$ & adventure, adventure-kids, family, etc.\\
	State - First Time User & $2$ & true/false\\
	State - Turn Depth & $110$ & $1, ..., 110$ \\
	\midrule
	Action & $8$ & offer-one-skill, offer-one-skill-or-category, \\
	& & offer-one-category, offer-three-categories, \\
	&& offer-five-categories, launch-skill, \\
	&& end-session, execute\\
 \bottomrule
\end{tabularx}
\label{table:spaces}
\end{table*}

\vspace{2pt}
\noindent
\textbf{Policy}: We use DQN \cite{Mnih2013,Mnih2015} with action masking for the RL agent, 
with a fully-connected MLP to represent the deep Q-network. 
The hidden layers use a rectifier nonlinearity, and the output layer is a 
fully connected layer with linear activation function and a single output for each valid action. 
The action mask suppresses impossible actions in any particular dialog state, 
such as launching a skill before the user has selected one. 

\section{Experimental Results}

We focused on the use case of a user searching for a game to play among $1,903$ skills belonging to $48$ game categories. Each category may also have subcategories, resulting in $191$ total categories.
Example of categories are adventure, trivia, choose your own story, family, and kids.
The number of categories to offer \textit{k} is set to one, three, and five; and the number of skills to offer \textit{n} is set to one, based on the results of internal user studies. Table \ref{table:spaces} summarizes the state and action spaces.
For all agents, we randomly sample from the set of possible prompts and metadata for the selected action. Furthermore, we used the Alexa Skill portal to train the NLU model from a set of sample utterances. 

\subsection{Simulation Results}
\label{rl-details}

\begin{figure}[!b]
\centering
    \caption{Success rate of DQN during training. The rule-based agent achieves an average success rate of $68.00\% (\pm 2\%$) in simulation.}
  		\includegraphics[width=0.58\textwidth]{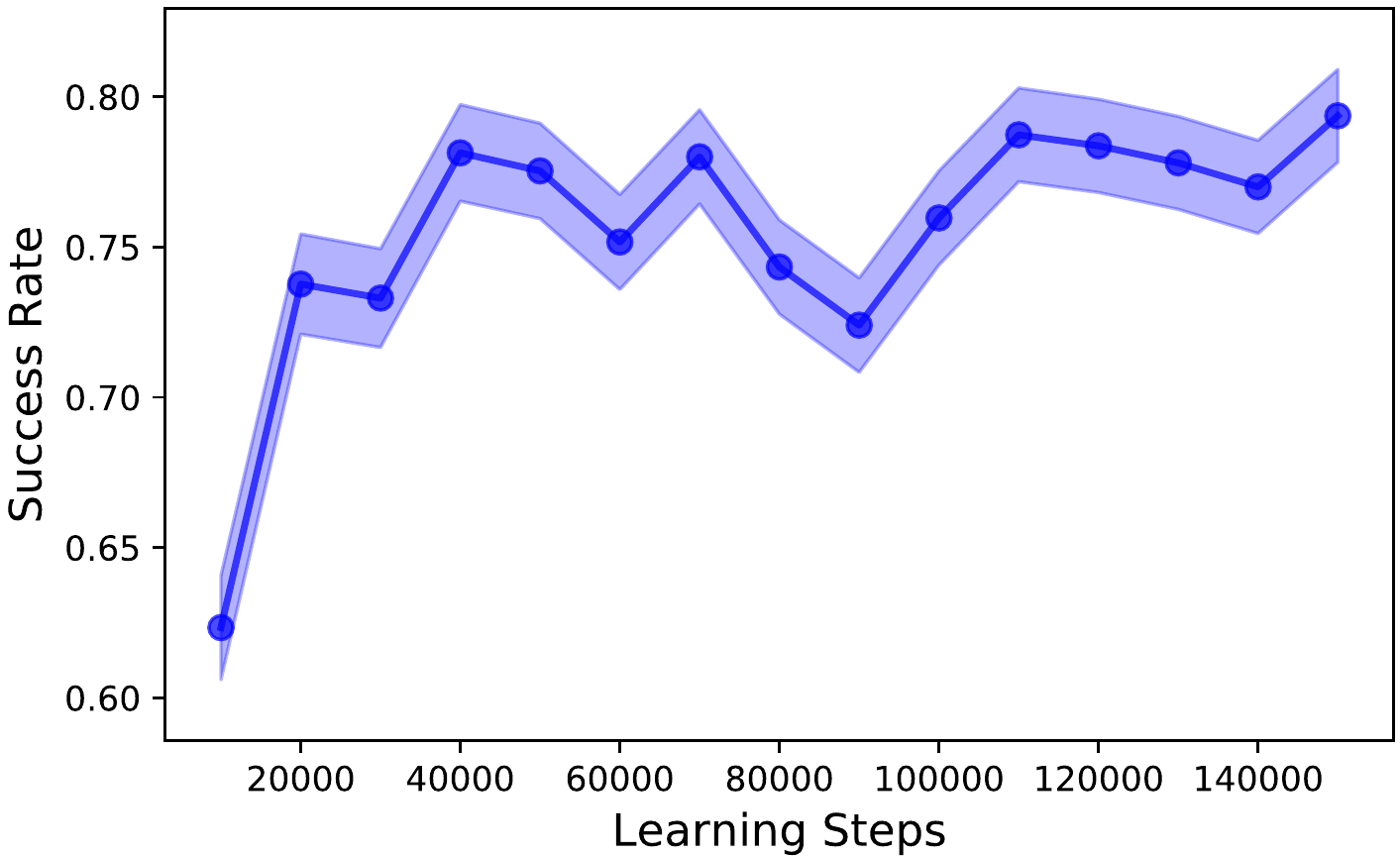}
  \label{learning_plots}
\end{figure}

We trained the DQN agent using an $\epsilon$-greedy policy with $\epsilon$ decreasing linearly from $1$ to $0.1$ over $100,000$ steps. 
Additionally, we tuned a window size to include previous dialog turns as input and set $\gamma$ to $0.9$. 
We ran the method $30$ times for $150,000$ steps, and in each run, after every 10,000 steps, 
we sampled $3,000$ dialog episodes with no exploration to evaluate the performance.
The optimal parameters were found using Hyperopt \cite{hyperopt} (see Appendix B).
Figure \ref{learning_plots} shows the simulation results during training. 
The Y-axis in the figure is the success rate of the agent (measured in terms of number of dialogs that resulted in launching a skill divided by total number of dialogs), and the X-axis is the number of learning steps. 
Given our choice of reward function, the increase in success rate is indicative of the agent learning to improve its policy over time.
Furthermore, the RL agent outperformed the rule-based agent with average success rate of $68.00\% (\pm 2\%$) in simulation. 

\subsection{Human Evaluation}

To evaluate the performance of the skill discovery agent, we deployed the dialog policies and evaluated them with real users (see Appendix A for examples of dialogs).
We first conducted a test with a baseline policy of recommending up to five skills
based on popularity and allowing the user to either accept or reject the recommendation. The success rate of this simple policy was $46.42$\%, illustrating the importance of providing flexible search to the user.
We then conducted an A/B test on the rule-based and RL policies to compare their effects on skill launches in a production environment. 
Both policies were tested on randomly sampled users, with the additional constraint of using the same policy for returning users.
The results are reported in Table \ref{table:ab_testing}. 
Both policies significantly outperform the baseline policy, indicating the importance of providing flexible search and navigation to users. 
Additionally, the difference between the success rate of the rule-based ($73.41$\%) and RL ($76.99$\%) policies is statistically significant ($p$-value $<$ $0.0001$) and the RL policy has significantly shorter dialogs ($p$-value $<$ $0.0001$), showing the importance of optimizing for the entire interaction with the user.

\begin{table}[!b]
 \centering
  \caption{Success rate and average dialog length for the rule-based and RL dialog policies.}
 \begin{tabular}{lccc}
 \toprule
   \textbf{Policy} & \textbf{Number} & \textbf{Success} & \textbf{Avg. Dialog}\\
 & \textbf{of Users}  & \textbf{Rate} & \textbf{Length}\\
 \midrule
	Rule-based& $32,032$ & $73.41\%$ & $4.97$\\
	\midrule
	RL-based & $32,295$ & $76.99\%$ & $4.65$\\
 \bottomrule
 \end{tabular}
 \label{table:ab_testing}
\end{table}

\begin{table}[!b]
 \centering
  \caption{Success rate of first-time and return users for the rule-based and RL dialog policies.}
 \begin{tabular}{lcc}
 \toprule
   \textbf{Policy} & \textbf{First-Time} & \textbf{Returning} \\
 \midrule
	Rule-based& $72.68\%$ & $74.49\%$ \\
	\midrule
	RL-based & $77.14\%$ & $76.76\%$ \\ 
 \bottomrule
 \end{tabular}
 \label{table:first_time}
\end{table}

In order to understand the effect of adapting to user attributes, 
we investigated the difference in success rate between first-time and returning users for the two policies. 
First-time users make up $59.48\%$ and $60.14\%$ of the population for the rule-based and RL policies, respectively.
Table \ref{table:first_time} shows the results.
The RL policy significantly outperforms the rule-based policy for both first-time ($p$-value $<$ $0.0001$) and returning users ($p$-value $= 0.0002$), indicating that the RL model has learned and adapted to user attributes.
Additionally, the RL policy has a similar performance for both groups of users. The difference for the rule-based policy between the two groups, on the other hand, is significant ($p$-value $= 0.0010$), indicating that this policy is more tuned to returning users.
This highlights the difficulty of authoring personalized dialog policies with rules, and shows the advantage of using RL for this problem.

\section{Related Work}
Conversational search and recommendation, especially in the context of e-commerce, have been explored by researchers \cite{Christakopoulou2016,Radlinski2017,Aggarwal2018,Zhang2018,Sun2018}. 
\citet{Christakopoulou2016} introduced an interactive recommendation protocol and studied whether to ask absolute or relative questions when gathering user preferences.
Their dialog system collects like/dislike and pairwise comparison feedback from users, and does not include actions typically present in a dialog system \cite{Sun2018}.
\citet{Radlinski2017} proposed a theoretical framework for conversational search.
\citet{Kenter2017} framed the problem as a machine reading task and applied it to question answering.
\citet{Aggarwal2018} developed a RL-based conversational search assistant, in which state and action spaces are domain specific and may require a significant amount of time to develop. 
\citet{Sun2018} proposed a unified framework to integrate recommender and dialog systems, in which instead of just returning the top-ranking results for a given user query, the agent attempts to optimize for long term reward by asking the user for the value of an attribute. In their work, the action space is limited to two types of actions, namely, requesting for the value of an attribute or making a recommendation.
\citet{Zhang2018} proposed a multi-memory network architecture and applied it to search and recommendation in e-commerce. Compared to previous works, our formulation of the search problem is domain independent, accounts for user attributes and conversational preferences, and includes actions typically present in a dialog system. Additionally, whereas existing works have not been evaluated in a real production setting, we conduct experiments with real users at large scale.

\section{Conclusion}

In this paper, we introduced the problem of conversational skill discovery in speech-based assistants and presented an approach to enable users to find skills.
To this end, we started with a rule-based agent and improved it by using RL, 
enabling the agent to adapt to different user attributes and conversational styles.
We compared popularity based, rule-based and RL-based model conversational agents by deploying them in a real production setting and showed
that the RL agent learns to adapt its policy to achieve a higher success rate with shorter dialogs.
For future work, we plan to further personalize the dialog policy based on user attributes and conversational preferences, and investigate richer state representations.
Furthermore, we plan to explore the impact of evolving attributes and preferences on the learned policies.

\subsubsection*{Acknowledgments}

We would like to thank the Alexa Machine Learning Platform team for making the customer experiments possible.
We would also like to thank Jared Casale, Jason Pazis, Longshaokan Wang, and Spyros Matsoukas for their feedback and support. 

\section*{Appendix}
\renewcommand{\thesubsection}{\Alph{subsection}}

\subsection{Examples of Dialogs}

\subsubsubsection{Dialog with the Rule-based Agent}

\begin{table}[htbp]
\small
 \centering
 \begin{adjustbox}{max width=1.0\textwidth}
 \bgroup
 \begin{tabularx}{1.0\linewidth}{lX}
  \toprule
  \multicolumn{1}{c}{\textbf{Speaker}} & \multicolumn{1}{c}{\textbf{Utterance}} \\
  \midrule
  User & Let's play a game \\
  Agent & Okay, let's find you a game to play. One of the most played games this week is Would You Rather for Family. Do you want to try it, or pick a game category? \\
 User & No \\
 Agent & What are you in the mood for? I have mystery games, multiplayer games, family games, sports games, geography games, or you can say more categories. \\
 User & Mystery games \\
 Agent & Mystery games, okay! Here are a few. I have a popular game called Escape the Room. Do you want to try it? \\
 User & No \\
 Agent & I have a popular game called The Wayne Investigation. Do you want to try it? \\
 User & No \\
 Agent &Next, I have The Vortex, which is rated as 4.0 by more than 900 people. Would you like to try it? \\
 User & No \\
 Agent & Okay, I have Guess the Number. Would you like to try it? \\
 User & Stop \\
 \bottomrule
 \end{tabularx}
 \egroup
 \end{adjustbox}
\end{table}

\newpage
\subsubsubsection{Dialog with RL-based Agent}

\begin{table}[htbp]
\small
 \centering
 \begin{adjustbox}{max width=1.0\textwidth}
 \bgroup
 \begin{tabularx}{1.0\linewidth}{lX}
  \toprule
  \multicolumn{1}{c}{\textbf{Speaker}} & \multicolumn{1}{c}{\textbf{Utterance}} \\
  \midrule
  User & Let's play a game \\
  Agent & Okay, let's find you a game to play. One of the most played games this week is Would You Rather for Family. Do you want to try it, or pick a game category? \\
 User & Pick a game category \\
 Agent & I have game show games, role playing games, top games, TV and movie games, echo button games, or other categories. Which of these do you want? \\
 User & Echo button games \\
 Agent & Echo button games, okay! Here are a few. First, do you want to try Party Foul? \\
 User & No \\
 Agent & Next, would you like to give Hanagram a try? \\
 User & No \\
 Agent & Would you like to try Don't Cut That Wire!, or would you like to pick a game category? \\
 User & Pick a game category \\
 Agent & What kind of game would you like? I have sports games, puzzle games, guessing games, or you can say more categories. \\
 User & Guessing games \\
 Agent & Okay, guessing games. First, would you like to give Twenty Questions a try? \\
 User & What? \\
 Agent & I have a popular game called Twenty Questions. Do you want to try it? \\
 User & Yes \\
 Agent & Gotcha. Here's Twenty Questions. \\
 \bottomrule
 \end{tabularx}
 \egroup
 \end{adjustbox}
\end{table}

\subsection{Learning Parameters for the RL Agent}

\begin{table}[H]
\small
  \centering
  \label{parameters}
  \begin{tabular}{l|c}
    \toprule
    \multicolumn{1}{c|}{}  & \multicolumn{1}{c}{DQN} \\
    \midrule
	Number of Hidden Layers & $2$ \\
	Number of Hidden Nodes per Layer & $128$  \\
	Learning Rate & $1e-05$  \\	
	Dropout & $0.3$ \\
	Experience Replay Size & $15,000$ \\
	Window Size & $1$  \\
	Embedding Size & $1$  \\
	Target Model Update Interval & $13,000$  \\
    \bottomrule
  \end{tabular}
\end{table}

\bibliographystyle{named}
\bibliography{d3m_codi}

\end{document}